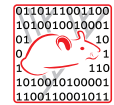

JOURNAL OF
BIOMEDICAL SEMANTICS

RESEARCH        Open Access

# OMG U got flu? Analysis of shared health messages for bio-surveillance

Nigel Collier[1,2*], Nguyen Truong Son[3], Ngoc Mai Nguyen[3]

*From* Fourth International Symposium on Semantic Mining in Biomedicine (SMBM)
Hinxton, UK. 25-26 October 2010

\* Correspondence: collier@nii.ac.jp
[1]National Institute of Informatics, 2-1-2 Hitotsubashi, Chiyoda-ku,Tokyo, Japan

## Abstract

**Background:** Micro-blogging services such as Twitter offer the potential to crowdsource epidemics in real-time. However, Twitter posts ('tweets') are often ambiguous and reactive to media trends. In order to ground user messages in epidemic response we focused on tracking reports of self-protective behaviour such as avoiding public gatherings or increased sanitation as the basis for further risk analysis.

**Results:** We created guidelines for tagging self protective behaviour based on Jones and Salathé (2009)'s behaviour response survey. Applying the guidelines to a corpus of 5283 Twitter messages related to influenza like illness showed a high level of inter-annotator agreement (kappa 0.86). We employed supervised learning using unigrams, bigrams and regular expressions as features with two supervised classifiers (SVM and Naive Bayes) to classify tweets into 4 self-reported protective behaviour categories plus a self-reported diagnosis. In addition to classification performance we report moderately strong Spearman's Rho correlation by comparing classifier output against WHO/NREVSS laboratory data for A(H1N1) in the USA during the 2009-2010 influenza season.

**Conclusions:** The study adds to evidence supporting a high degree of correlation between pre-diagnostic social media signals and diagnostic influenza case data, pointing the way towards low cost sensor networks. We believe that the signals we have modelled may be applicable to a wide range of diseases.

## Introduction

Rising awareness of infectious disease outbreaks and the high costs of extending traditional sensor networks means that we have an opportunity to harness new forms of social communication for crisis surveillance. The trend is already underway with automatic map generation from Twitter reports for earthquakes and typhoons [1,2], the symptom-based influenza tracking portal Flutracking [3] as well as the humanitarian portal Ushahidi [4]. Despite a risk of high false reporting rates there is nevertheless strong potential in having multiple sensor sources for verification, robustness and redundancy. In the case of earthquake detection, Earle notes that Twitter messages (tweets) can be available up to 20 minutes before the official report from the US Geological Survey. With epidemics too the time period from signal to detection is critical.

Recent studies such as [5] estimate that the average delay in receiving and disseminating data from traditional sentinel physician networks is about two weeks.

A small but growing number of early warning systems have already developed to mine event information from low cost Web sources mainly focusing on edited newswire reports (see Hartley *et al.*[6] for a survey). Success in operationalizing such systems has crucially depended on building close collaborations with government and international public health agencies in order to perform detailed verification and risk assessment.

Recent studies on alerting from newswire reports [7] are beginning to make clear the operational boundaries in terms of their selectivity, volume and timeliness. In this earlier work the first author (NC) noted the issue of late warnings, i.e. where there is a known outbreak in a country but spikes in true alerts at the province or city level are occluded by low rising aggregated data for the country as a whole. To overcome such problems micro-blogging might have a role to play. Micro-blogs may be able to help also with very early epidemic detection, i.e. at the pre-diagnostic stage where there is maximum scientific uncertainty about symptoms, transmission routes and infectivity rates. Automatic geo-coding and the ability to send messages from many types of mobile device are a key advantage in this respect.

## Background

In micro-blogging services such as Twitter, users describe their experiences directly in near-real time in short 140 character tweets. As of April 2010 it was estimated that Twitter had approximately 106 million registered users with 300,000 new users being added each month. Despite their potential coverage, timeliness and low overhead, tweets present their own unique challenges: pre-diagnostic unedited reports mean that there is a large trust issue to resolve within the modeling technique. Social media can also reflect a high degree of reactivity to risk perception as seen during the H1N1 pandemic in 2009 – redistributing links or requests for information rather than generating user experience. To an extend this reflects newswire coverage and the amount of uncertainty readers feel. Re-tweets in themselves may provide useful signal but their role has yet to be quantified. Despite these obvious challenges we believe there is potential for using very short messages to detect epidemic trends, as hinted at by the success of Google Flu Trends [8] which harness user's search queries.

In order to do this we propose to employ aberration detection for detecting sharp rises in the features that signal epidemics. A precursor to this is in identifying reliable features themselves. In this study we started by looking at reports of individual experience on Twitter with a focus on the precautionary actions as identified by Jones and Salathé in their behaviour response survey [9] for A(H1N1). Modeling individual risk perception based on local health information appears to be an understudied area in Web-based health alerting which may add signal to event-based early detection models.

Recently a number of studies have appeared looking at the effectiveness of search queries and social media. Lampos *et al.*[10] studied tweets in the 49 most populated urban centres of the UK and found a strong linear correlation with Health Protection Agency influenza like illness (ILI) data from general practitioner (GP) consultations during the 2009-2010 influenza season. Studies on user query data from Google Flu Trends has also shown strong correlations with sentinel network data. Valdivia *et al.*

[11] showed for the 2009 Influenza A(H1N1) pandemic there was a strong Spearman's Rho correlation with ILI and acute respiratory infection (ARI) data from sentinel networks in Europe.

Nevertheless challenges in interpreting query and social networking data remain. Ortiz *et al.*[12] for example discuss the potential for confusion in Google Flu Trends between ILI and non-influenza illnesses. Influenza data was compared from Google Flu, the CDC outpatient surveillance network and the US influenza virological surveillance system. Whilst correlation with ILI was found to be high, it was found that correlation with actual influenza test positive results was lower. This result highlights the fact that both social media and user queries are secondary indicators that should be correlated with patient reported symptoms. Significant deviation between user's searching behaviour and ILI rates was noted for the 2003-2004 influenza season when influenza activity, pediatric deaths and news media coverage of influenza were particularly high. This highlights another understudied issue: that we need to work hard to remove elements of reporting bias during media storms.

## Methods
### Annotation
Taking Jones and Salathé's behaviour responses as a starting point we surveyed potential messages in Twitter in relation to H1N1 influenza topics. Prom an initial group of thirteen categories we decided, due to low frequency counts, to conflate several into a final grouping of four. e.g. avoiding people who cough/sneeze, avoiding large gatherings of people, avoiding school/work and avoiding travel to infected areas were joined into a general 'avoidance behaviour' category. To this we added a final category for direct reporting of influenza. The final list of categories is: (A) Avoidance Behavior – behaviours which avoid agents thought to be at risk of infection; (I) Increased sanitation – sanitation measures to promote individual health and prevent infection; (P) Seeking pharmaceutical intervention – seeking clinical advice or using medicine or vaccines; (W) Wearing a mask; and (S) Self reported diagnosis – reporting that one has influenza. The first four categories correspond approximately to Bish and Michie's [13] three broad categories of protective behaviour: preventive (I, W and vaccinations from P), avoidance (A), and management of disease (using medication and seeking clinical advice from P but excluding using the Internet and help lines).

As expected there are a number of caveats to each of these broad classes. We list up only a representative sample here: (1) A message is only tagged positive if the user or a close family member is the subject of the tweet; (2) If the message indicates that the action is hypothetical then the classification is negative; (3) The time of the reported event should be within one week of the current time; (4) Messages can belong to more than one category. Examples of (anonymised) messages are shown in Table 1.

At a practical level the problem of identifying self protection messages can be characterised as classifying very biased data. In order to handle this we adopted two stages of filtering. The first stage used a bag of 7 keywords to select tweets on topics related to influenza (*flu, influenza, H1N1, H5N1, swine flu, pandemic, bird flu*). For 1st March 2010 to April 30th 2010 this resulted in a pool of about 225,000 tweets. This first stage of filtering was also designed to reduce the ambiguity of keywords such as 'fever' and 'cough' which occur in a wide variety of contexts. For example 'cough' is

**Table 1 Example messages**

| n | Message | A | I | P | W | S |
|---|---|---|---|---|---|---|
| e1 | home this weekend? i've been off work all week with the flu | + | - | - | - | + |
| e2 | there is alot more to preparing for Swine Flu than just washing your hands | - | - | - | - | - |
| e3 | everyone wash your hands.. no one wants swine flu | - | + | - | - | - |
| e4 | awl u need to go get to the doc so u dnt past da swine flu | - | - | - | - | - |
| e5 | it's 2:10pm, I have flu and I'm still wearing my pajama | - | - | - | - | + |
| e6 | I have the flu. I had a normal flu shot | - | - | + | - | + |
| e7 | This guy has a nasty cough! Thank god he's sitting far away from me - the swine flu travels | + | - | - | - | - |
| e8 | I'm sick too… cold or flu, I don't know… I couldn't go to work today… | + | - | - | - | + |
| e9 | Trivia for tonight has been cancelled due to flu bug | + | - | - | - | - |
| e10 | Feel like I've washed my hands a 1000 times Gotta loveworkin during cold & flu season | - | + | - | - | - |
| e11 | overhyped public scare. I want to remove this mask | - | - | - | + | - |
| e12 | i don't know. she just keeps getting sick, but it's not the flu. i hate keeping her off school | - | - | - | - | - |
| e13 | i feel terrible, don't want to be at work, wish id never had the h1n1 jab | - | - | + | - | - |
| e14 | Some cleaning products were especially made to kill the H1N1 … | - | - | - | - | - |
| e15 | She has a surgical mask on in the movies I'm nervous hope it's not h1n1 | - | - | - | - | - |
| e16 | regretting not getting a flu shot this year | - | - | - | - | - |

Positive (+) and negative (-) examples of classified messages.

frequently used in tweets to indicate irony, e.g. *that's true –***cough cough,** and 'fever' often indicates intense enthusiasm, e.g. *Christmas fever, Bieber fever, app fever,* etc.

The second stage used hand built patterns to select a total of 14,508 tweets. From these we randomly chose 7,412 tweets spread across the five classes. All duplicates were removed leaving 5,283 messages and the resulting data was then classified by hand using a single annotator as detailed in Table 2. Results for mean character length and standard deviation showed no category-specific trend except to illustrate the wide variety of message lengths.

In order to test the stability of the annotation scheme and our assumptions about its reproducibility we calculated kappa for 2,116 messages balanced across all the classes. For this another annotator was chosen who did not take part in the creation of the guidelines and was not a co-investigator in this study. The simple agreement ratio was 0.88 (the total number of matched class assignments divided by the total number of messages). Kappa was calculated as $\kappa = (pA - pE)/(1 - pE)$, where $pA$ was 0.88 and $pE$ was 0.12. $\kappa$ was then found to be 0.86. Both results reveal a high level of agreement in the annotation scheme and give us confidence to move ahead with automated classification. Note that the final annotation scheme guidelines can be obtained on request from the first author.

**Table 2 Message frequency by class**

|  | A | I | P | W | S |
|---|---|---|---|---|---|
| Positive | 251 | 37 | 499 | 32 | 741 |
| Negative | 632 | 43 | 974 | 230 | 1873 |
| Total | 883 | 80 | 1443 | 262 | 2614 |
| Mean length | 109.2 | 118.8 | 107.0 | 117.3 | 100.9 |
| Sd. length | 28.9 | 21.9 | 30.6 | 27.7 | 33.4 |
| N/P ratio: | 0.40 | 0.86 | 0.51 | 0.14 | 0.40 |

Message frequency in the training/testing corpus for self-protection classes. Mean message length, standard deviation for message length and the ratio of negative to positive messages are also shown.

### Models

We employed two widely used classification models implemented in the Weka Toolkit [14], Naive Bayes (NB) and Support Vector Machines (SVMs) [15] classifying five data sets (total 5283 messages) into positive or negative. SVM used a RBF-kernel and grid search for finding the best parameter settings. Since we hypothesized that custom built regular expressions might have more traction for achieving high precision we decided to use a freely available toolkit called the Simple Rule Language (SRL) [16] for this purpose. SRL comes with an interface for maintaining the rule base which can be run in testing mode to convert surface expressions into structured information.

SRL rules were built from a held out set of tweets not used in training. Rules consist of string literals, skip expressions, word lists, named entity classes and guard expressions for limiting the scope of matched entities. Rule building took approximately 10 hours of work. The rule book contains specialised synonym sets to recognize common and slang terms for medicines (e.g. *shot, vaccine, drug, tamiflu, jab, medicine, vacc*), physicians (e.g. *doctor, doc, dr, physician*) and other key domain entities. Verb lists are maintained for specialized lexical classes such as prescribe (e.g. *prescribe, perscribe\**). Lists are also built for pronouns, common temporal adverbs, modal verbs and negations. Special rules were built to recognize past events. The exceptional class was I (increased sanitation) where we were not able to identify enough examples with confidence to build meaningful rules by hand. In this case only unigrams and bigrams were used to train the classifiers.

We found that the language used in tweets to express user's behaviour is very diverse and idiosyncratic so it is challenging to achieve a high degree of coverage in the rules with surety. With this in mind we combined features from the rules with unigrams and bigrams. If a rule matched a tweet its feature value was set to 1, otherwise to 0.

### Results 1: classification experiments

Test runs used 10-fold cross validation on categories A, P and S and 3-fold cross validation on categories I and W due to small number of items. We calculated recall, precision and F-score performance for each category. As we can see by comparing the results in Tables 3 and 4, SVM overall performs better on all categories except for W (wearing a mask). The low F-scores for the W category seem to be a result of the strongly biased negative data in this category, conforming to predictions of earlier studies on imbalanced data sets such as [17]. To confirm this assumption we conducted undersampling on the negative data so that the number of items equaled the number of positive items within each class. We then repeated the experiments and found that overall, performance was uniformly higher and that classes like W which performed poorly did well (data not shown).

Both SVM and NB model's performance generally follows the amount of training data except for S (Self diagnosis) where the F-score is slightly lower than the trend in other classes despite large numbers of positive examples. The overall trend for NB is to have stronger recall than precision whereas for SVM precision is generally higher than recall.

The results suggest that our SRL rule book seemed to offer substantial benefits when combined with unigrams but less certain improvements when combined with unigrams plus bigrams. Looking slightly deeper into the results we found a correlation between

**Table 3 Results for Naive Bayes classification**

|  | P | R | F1 |
|---|---|---|---|
| **A** | | | |
| UNI | 0.55 | 0.77 | 0.64 |
| UNI+SRL | 0.56 | 0.80 | 0.66 |
| UNI+BI | 0.54 | 0.80 | 0.65 |
| UNI+BI+SRL | 0.56 | 0.80 | 0.66 |
| **I** | | | |
| UNI | 0.54 | 0.57 | 0.55 |
| UNI+BI | 0.48 | 0.43 | 0.46 |
| **P** | | | |
| UNI | 0.60 | 0.80 | 0.68 |
| UNI+SRL | 0.61 | 0.81 | 0.70 |
| UNI+BI | 0.61 | 0.83 | 0.70 |
| UNI+BI+SRL | 0.62 | 0.84 | 0.71 |
| **W** | | | |
| UNI | 0.24 | 0.63 | 0.35 |
| UNI+SRL | 0.29 | 0.78 | 0.42 |
| UNI+BI | 0.25 | 0.72 | 0.37 |
| UNI+BI+SRL | 0.26 | 0.72 | 0.38 |
| **S** | | | |
| UNI | 0.53 | 0.70 | 0.61 |
| UNI+SRL | 0.59 | 0.74 | 0.65 |
| UNI+BI | 0.54 | 0.78 | 0.64 |
| UNI+BI+SRL | 0.57 | 0.78 | 0.66 |

F1 results for tweet classification using Naive Bayes. UNI = unigram, BI = bigram, SRL = Simple Rule Language regular expression.

message length and classification accuracy in NB and SVM. For NB, whilst the length of messages didn't seem to make much difference to the false negative rate which remained constant at about 0.2 to 0.25 on messages in the length range of 34 to 144 characters, it impacted to a greater degree on false positives (0.23 on shorter messages of length 34 to 56 down to 0.08 for messages of length 122 to 144). For SVM there appeared to be a general reduction in both false positives and false negatives as message length increased.

As expected, frequent misspellings, abbreviated word forms, slang and lack of punctuation complicated the classification task. Missing auxiliary verbs and articles need to be compensated for within the SRL rules in order to ensure successful matching.

### Results 2: comparison to CDC data

In order to provide a proof of concept we operationalised the classifiers and ran them on a corpus of Twitter data called the Edinburgh Corpus [18]. The Edinburgh corpus holds 97 million tweets for the period November 11th 2009 to February 1st 2010 from 9 million users. This represents over 2 billion words from a variety of languages. Of these 12.5 million are reported as topic tags, 55 million are @ replies and 20 million are links.

We applied the same keyword filtering method used on the Edinburgh corpus for the first set of experiments and obtained 52,193 tweets for the period of study. Following this we applied the SVM UNI model and then compared the week by week volumes

**Table 4 Results for SVM classification**

|  | P | R | F1 |
|---|---|---|---|
| A |  |  |  |
| UNI | 0.70 | 0.66 | 0.68 |
| UNI+SRL | 0.72 | 0.69 | 0.70 |
| UNI+BI | 0.71 | 0.70 | 0.70 |
| UNI+BI+SRL | 0.71 | 0.71 | 0.71 |
| I |  |  |  |
| UNI | 0.62 | 0.70 | 0.66 |
| UNI+BI | 0.61 | 0.59 | 0.60 |
| P |  |  |  |
| UNI | 0.65 | 0.84 | 0.73 |
| UNI+SRL | 0.65 | 0.85 | 0.74 |
| UNI+BI | 0.67 | 0.77 | 0.72 |
| UNI+BI+SRL | 0.67 | 0.78 | 0.72 |
| W |  |  |  |
| UNI | 0.15 | 0.06 | 0.09 |
| UNI+SRL | 0.25 | 0.16 | 0.19 |
| UNI+BI | 0.15 | 0.06 | 0.09 |
| UNI+BI+SRL | 0.31 | 0.16 | 0.21 |
| S |  |  |  |
| UNI | 0.64 | 0.59 | 0.61 |
| UNI+SRL | 0.68 | 0.72 | 0.70 |
| UNI+BI | 0.74 | 0.54 | 0.63 |
| UNI+BI+SRL | 0.78 | 0.60 | 0.67 |

F1 results for tweet classification using SVM. UNI = unigram, BI = bigram, SRL = Simple Rule Language regular expression against laboratory results for weeks 47 to 5 of the 2009-2010 influenza season in the USA [19]. Counts are shown in Table 5. Several interesting trends can be observed: (a) The total volume of positively identified Tweets was relatively small compared to the volume of Tweets as a whole; (b) Wearing a mask was totally absent from our classified data; (c) The aggregated counts for self protection (A+I+P, data not shown) seem to have a close correlation to CDC results (data not shown). To measure correlation

**Table 5 Twitter positives versus CDC cases**

| Wk | A | S | I | P | CDC |
|---|---|---|---|---|---|
| 46[A] | 49 | 48 | 22 | 222 | 2715 |
| 47 | 32 | 72 | 30 | 258 | 1408 |
| 48 | 24 | 49 | 9 | 181 | 997 |
| 49 | 35 | 41 | 10 | 199 | 610 |
| 50 | 35 | 39 | 10 | 154 | 480 |
| 51 | 21 | 35 | 12 | 150 | 251 |
| 52 | 19 | 26 | 4 | 37 | 285 |
| 1 | 25 | 32 | 6 | 63 | 266 |
| 2 | 25 | 32 | 5 | 81 | 261 |
| 3 | 29 | 31 | 7 | 73 | 317 |
| 4 | 29 | 20 | 7 | 62 | 268 |
| 5 | 29 | 23 | 6 | 46 | 290 |

Positively identified Tweets in the Edinburgh corpus shown against Influenza Positive tests reported to CDC by U.S. WHO/NREVSS collaborating laboratories, National Summary, 2009-2010. Counts for W were zero throughout and are therefore not shown. [A] For week 46 we only have partial Twitter data available in the Edinburgh corpus.

we calculated the Spearman's Rho (see http://en.wikipedia.org/wiki/Spearman% 27s_rank_correlation_coefficient) between counts of positive messages in each class and the CDC laboratory data for A(H1N1). Table 6 shows moderately strong correlations. The strongest correlation appeared when A,I and P were combined. Besides W which failed to provide any data, the weakest evidence came from Increased sanitation (I). Differences could be due to (a) the global geographic coverage of tweets in our collection; and (b) the syndromes covered in our self protection behaviour and self reporting messages are wider than A(H1N1) and could actually be other diseases such as common colds, strep throat, adenovirus infection and so on.

Drill down analysis reveals that we still need to do more to remove false positives by strengthening the linguistic features within the limits of the 140 character length. Examples of false positives include interrogative sentences, hypothetical sentences, reports on events that took place in the distant past, comments on influenza advice from others, etc.

## Conclusions

In this paper we have made the first steps towards classifying Twitter messages according to self reported risk behaviour. The results have shown moderately strong correlation with CDC A(H1N1) data but we still need to make further progress in order to achieve the high degrees of correlation reported between Google Flu trends and sentinel influenza data. The next step will be to extend our training data, automatically compensate for imbalanced classes, strengthen the linguistic features and see if we can use these signals to detect emerging disease outbreaks. It was shown in Jones and Salathé that after an initial peak in levels of risk concern, anxiety faded once the immediate threat of the A(H1N1) pandemic had passed. In follow up work with data over longer time periods we intend to look at how closely these signals track epidemic case data.

In terms of the limitations of this study, we found that the volume of positively classified messages was lower than we had expected and we could not yet find evidence that pre-diagnostic self-protective or self-reporting messages were predicting rises in the case data from laboratories. The effects of reporting bias were largely ignored in our study and future work will need to take account of the demographic characteristics of Twitter users in terms of their specific location and average age. For example it has been found by Lau *et al.*[20] that older people are more likely to report self-protective behaviour in an avian influenza pandemic. We also ignored possible associations between preventive behaviours and the perceived severity of the disease [13].

**Table 6 Correlation between Twitter positives and CDC cases**

| Category | Spearman's Rho[a] | p-value[b] |
|---|---|---|
| A | 0.66 | 0.020 |
| S | 0.66 | 0.021 |
| I | 0.58 | 0.048 |
| P | 0.67 | 0.017 |
| A+I+P | 0.68 | 0.008 |
| A+I+P+S | 0.67 | 0.017 |

Correlation between CDC AH1N1 laboratory data frequency for Influenza 2009-2010 and aggregated self protection behaviour counts and self reported diagnosis from Tweets. [a] Spearman's rank-order correlation coefficient. [b] p values are reported for a two-tailed test. Calculations were done using VassarStats (http://faculty.vassar.edu/lowry/corr_rank.html)

We believe that the signals we have modeled may be applicable to a wide range of diseases and we intend to explore how these features can be used to detect diseases other than influenza. Besides disaster alerting, results from analysis of behavioural responses may also help in the future to evaluate the success of official prevention campaigns. For example, it is known that little notice was taken of antiviral therapies, goggle or mask wearing advice in the Netherlands after the Avian Influenza epidemic was introduced to Europe [21]. Conversely, empirical studies of individual risk perception in disease severity and susceptibility may help official agencies to tune risk communication strategies more effectively.

**List of abbreviations**
GP: General Practitioner; ILI: Influenza like illness; NB: Naive Bayes; SVM: Support Vector Machines; SRL: Simple Rule Language; CDC: United States Centers for Disease Control and Prevention

**Acknowledgements**
We gratefully acknowledge grant support from the National Institute of Informatics for this study. The research work in its first unrevised form was presented at the SMBM 2010, Hinxton, Cambridge, U.K.
This article has been published as part of *Journal of Biomedical Semantics* Volume 2 Supplement 5, 2011: Proceedings of the Fourth International Symposium on Semantic Mining in Biomedicine (SMBM). The full contents of the supplement are available online at http://www.jbiomedsem.com/supplements/2/S5.

**Author details**
[1]National Institute of Informatics, 2-1-2 Hitotsubashi, Chiyoda-ku,Tokyo, Japan. [2]Japan Science and Technology Agency, 2-1-2 Hitotsubashi, Chiyoda-ku,Tokyo, Japan. [3]Vietnam National University at HCMC, Ho Chi Minh City, Vietnam.

**Author contributions**
The study was conceived and directed by NC, corpus collection, analysis and guideline development was done by NM and the machine learning experiments were done by NS. All authors contributed and agreed to the final version of this manuscript.

**Competing interests**
The authors declare that they have no competing interests.

Published: 6 October 2011